\documentclass[sigconf]{acmart}
\usepackage{booktabs}
\usepackage{subfigure}
\usepackage{enumitem,bm}
\usepackage{rotating}
\usepackage{amsmath,amssymb,amsfonts}

\usepackage{multirow}
\usepackage{booktabs}
\usepackage{algorithm}
\usepackage{algorithmic}

\AtBeginDocument{%
  \providecommand\BibTeX{{%
    \normalfont B\kern-0.5em{\scshape i\kern-0.25em b}\kern-0.8em\TeX}}}

\copyrightyear{2023}
\acmYear{2023}
\setcopyright{acmlicensed}
\acmConference[MM '23] {Proceedings of the 31st ACM International Conference on Multimedia}{October 29--November 3, 2023}{Ottawa, ON, Canada.}
\acmBooktitle{Proceedings of the 31st ACM International Conference on Multimedia (MM '23), October 29--November 3, 2023, Ottawa, ON, Canada}
\acmPrice{15.00}
\acmISBN{979-8-4007-0108-5/23/10}
\acmDOI{10.1145/3581783.3612026}
\def\method{ALEX}

\begin{document}

\title[\method{}]{\method{}: Towards Effective Graph Transfer Learning \\with Noisy Labels}

\def\affil{National Key Laboratory for Multimedia Information Processing, \\School of Computer Science, \\ Peking University}

\author{Jingyang Yuan}
\orcid{0009-0001-0831-5600}
\affiliation{%
  \institution{\affil{}}
  \city{Beijing}
  \country{China}
}
\email{yuanjy@pku.edu.cn}

\author{Xiao Luo}
\orcid{0000-0002-7987-3714}
\affiliation{%
  \institution{Department of Computer Science, University of California, Los Angeles}
  \city{Los Angeles}
  \country{USA}}
\email{xiaoluo@cs.ucla.edu}

\author{Yifang Qin}
\orcid{0000-0002-7520-8039}
\affiliation{%
  \institution{\affil{}}
  \city{Beijing}
  \country{China}
}
\email{qinyifang@pku.edu.cn}

\author{Zhengyang Mao}
\orcid{0000-0002-2277-6008}
\affiliation{%
  \institution{\affil{}}
  \city{Beijing}
  \country{China}
}
\email{zhengyang.mao@stu.pku.edu.cn}

\author{Wei Ju}
\authornote{Corresponding authors.}
\orcid{0000-0001-9657-951X}
\affiliation{%
  \institution{\affil{}}
  \city{Beijing}
  \country{China}
}
\email{juwei@pku.edu.cn}

\author{Ming Zhang}
\authornotemark[1]
\orcid{0000-0002-9809-3430}
\affiliation{%
  \institution{\affil{}}
  \city{Beijing}
  \country{China}
}
\email{mzhang_cs@pku.edu.cn}

\renewcommand{\shortauthors}{Jingyang Yuan et al.}



\begin{abstract}
Graph Neural Networks (GNNs) have garnered considerable interest due to their exceptional performance in a wide range of graph machine learning tasks. Nevertheless, the majority of GNN-based approaches have been examined using well-annotated benchmark datasets, leading to suboptimal performance in real-world graph learning scenarios. To bridge this gap, the present paper investigates the problem of graph transfer learning in the presence of label noise, which transfers knowledge from a noisy source graph to an unlabeled target graph. We introduce a novel technique termed B\underline{a}lance A\underline{l}ignment and Information-awar\underline{e} E\underline{x}amination (\method{}) to address this challenge.
\method{} first employs singular value decomposition to generate different views with crucial structural semantics, which help provide robust node representations using graph contrastive learning. To mitigate both label shift and domain shift, we estimate a prior distribution to build subgraphs with balanced label distributions. Building on this foundation, an adversarial domain discriminator is incorporated for the implicit domain alignment of complex multi-modal distributions. Furthermore, we project node representations into a different space, optimizing the mutual information between the projected features and labels. Subsequently, the inconsistency of similarity structures is evaluated to identify noisy samples with potential overfitting. Comprehensive experiments on various benchmark datasets substantiate the outstanding superiority of the proposed \method{} in different settings.

\end{abstract}

%

\maketitle

\section{Introduction}

Graphs have gained considerable prominence in both contemporary research and practical applications, encompassing areas such as social network analysis~\cite{zhang2022improving,salamat2021heterographrec,fan2019graph,ju2023tgnn}, multimedia analysis~\cite{luo2020adversarial,wang2019zero,wang2021video,ji2022multimodal}, recommendation systems~\cite{fan2019graph,wang2021dualgnn,fan2020graph,cai2023lightgcl,ju2022kernel} and among others. As the development of deep learning, graph neural networks (GNNs) have emerged as a popular approach for learning effective representations of structured data~\cite{kipf2017semi,hamilton2017inductive,velivckovic2018graph}, demonstrating exceptional performance across a wide array of graph machine learning challenges. The core principle of GNNs involves implementing the message passing mechanism, where each node refines its representation by obtaining and aggregating information from its neighboring nodes.

\begin{figure}[]
\centering
\includegraphics[width=0.5\textwidth,keepaspectratio=true]{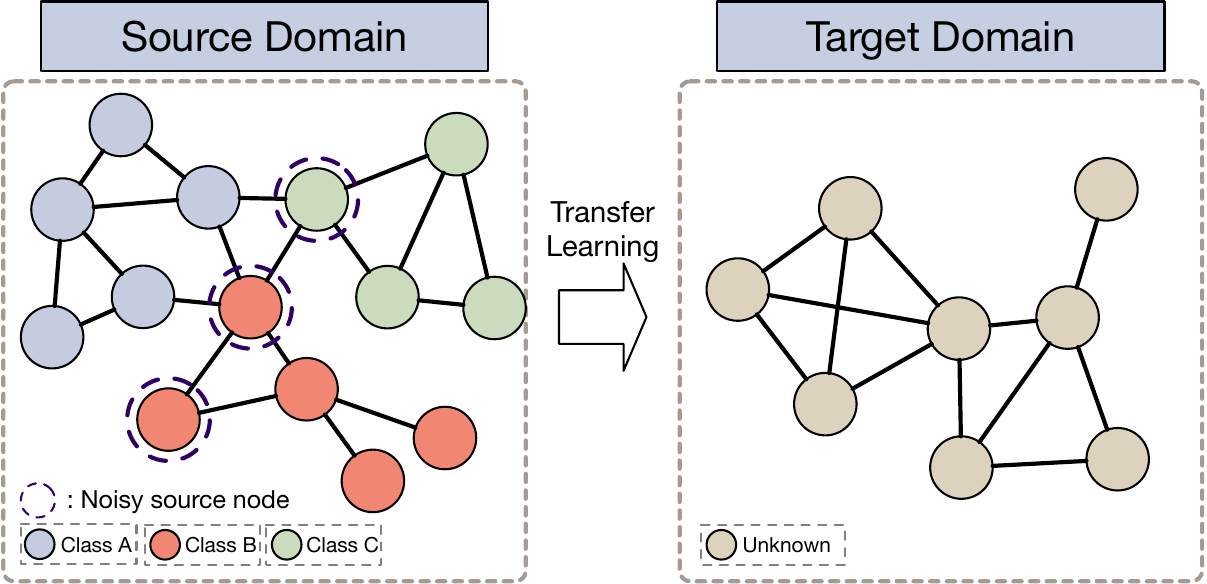}
\caption{The problem setting of graph transfer learning under label noise. We aim to transfer knowledge from a source graph with label noise to an unlabeled target graph. 
}
\label{fig:def}
\end{figure}

Despite their remarkable performance, current GNN-based methods are predominantly evaluated using standard benchmark datasets, which may encounter two challenges as follows when applied to real-world situations: (1) \textbf{Significant domain shift}: Contemporary methods~\cite{kipf2017semi,hamilton2017inductive,velivckovic2018graph} often focus on transductive learning within a single graph, whereas real-world applications could involve numerous online graphs~\cite{wu2020unsupervised}. However, transferring knowledge from source graphs to target graphs remains a non-trivial problem necessitating the management of significant domain shifts. (2) \textbf{Potential noisy signals}: Typically, these methods erroneously assume that the semantic labels in the training sources are free from errors~\cite{dai2021nrgnn}. In reality, modern engineers employ machines to enhance annotation efficiency~\cite{gong2014multi,zhang2021dualgraph}, which could introduce label noise, leading to substantial overfitting and compromised generalization capabilities. In light of these challenges, the paper explores the practical problem of graph transfer learning under label noise, wherein GNNs are transferred from a source graph with potential label noise to an unlabeled target graph exhibiting severe domain shift.

As label noise and domain shift would have a mixed effect, it is challenging to construct a powerful GNN under these conditions. Firstly, extant graph transfer learning approaches~\cite{wu2020unsupervised,zhu2021transfer,zhang2021adversarial,shen2020network,wu2022attraction} often employ explicit or implicit strategies to align data distributions across various graphs. 
Nevertheless, considerable label noise can impede the accurate capture of underlying multi-modal distributions~\cite{long2018conditional} in graphs, rendering domain alignment arduous. Secondly, the majority of robust GNN methods are applied to semi-supervised settings wherein numerous unlabeled nodes are leveraged to supply additional semantic information~\cite{nt2019learning,dai2021nrgnn,qian2023robust}. However, a large number of nodes within the target graph could introduce potentially biased pseudo-labels due to significant domain discrepancies, leading to error accumulation.
Lastly, label distributions across two graphs could exhibit considerable differences, i.e., label shift~\cite{liu2021adversarial}, which could further exacerbate the already challenging task of aligning the underlying multi-modal distributions in graphs.

In this paper, we propose a novel graph neural network named B\underline{a}lance A\underline{l}ignment and Information-awar\underline{e} E\underline{x}amination (\method{}) for this real-world problem. 
To mitigate the overfitting of noisy nodes, our \method{} learns robust node representations utilizing graph contrastive learning. In particular, based on the homophily assumption of graphs~\cite{wang2022hagen}, singular value decomposition~\cite{baker2005singular} is employed to generate a low-rank adjacency matrix, providing different graph views while preserving essential structural information.
To surmount domain shift along with label shift, we estimate a prior distribution that generates subgraphs with balanced label distributions. Subsequently, we construct a domain discriminator conditioned on both node embeddings and label distributions, which are trained to align underlying multi-modal distributions behind graphs in an adversarial manner. Additionally, we identify potentially noisy nodes based on mutual information. Specifically, we project node representations into a different space in which the mutual information between their representations and labels is maximized. As graph structures are not incorporated, the embedding vectors of noisy nodes could be substantially influenced. Therefore, in order to eliminate label noise, we select nodes exhibiting high inconsistency in similarity structures for each node between the two spaces. To summarize, the contribution of this paper is as follows:
\begin{itemize}[leftmargin=*]
    \item \textbf{Problem:} This paper studies an underexplored yet realistic problem of graph transfer learning under label noise and proposes a unified method \method{} to tackle this. 
    \item \textbf{Methodology:} Our \method{} not only studies both prior distributions for balanced domain alignment of robust node representations but also estimates the inconsistency of similarity structures to detect noisy nodes after maximizing mutual information between representations and labels. 
    \item \textbf{Evaluation:} Extensive experiments on various datasets validate the superiority of \method{} compared with state-of-the-art baselines.
\end{itemize}

\section{Related Work}
\label{sec::related}

\subsection{Graph Neural Networks}

Due to their high capacity for modeling graph-structured data, graph neural networks (GNNs) are gaining popularity in many communities~\cite{zhang2022improving,salamat2021heterographrec,fan2019graph,ju2023comprehensive,luo2022clear,luo2023hope}. After simplification from spectral methods, the majority of graph neural networks adhere to the message passing principle~\cite{kipf2017semi}, where each node updates its representation by iteratively aggregating information from neighbors. There have been extensive efforts to enhance the effectiveness and efficacy of GNNs from multiple perspectives, including architecture design~\cite{xhonneux2020continuous,wu2020comprehensive}, trustworthy~\cite{wang2021confident,yuan2020xgnn} and invariance theory~\cite{chen2022learning,wu2022discovering}. For instance, GraphSAGE~\cite{hamilton2017inductive} accelerates the rate at which messages travel through sampling nodes in the vicinity surrounding each node. CIGA~\cite{chen2022learning} employs causal models to characterize the generation of graphs and enhances the generalization ability by invariant learning. However, these methods do not account for realistic scenarios involving label noise and domain shift, which could impact their practice use. Towards this end achieve this objective, we proposes a novel method \method{} for these realistic scenarios.

\subsection{Domain Adaptation}
As a popular topic in the multimedia community, domain adaptation seeks to convey semantic information from a labeled source domain to an unlabeled target domain~\cite{ijcai2022p232,ijcai2022p213,li2022domain}. Earlier attempts measured the distance between the source and target domains using different matrices, e.g., maximum mean discrepancy~\cite{chang2021unsupervised} (MMD) and enhanced transport distance~\cite{li2020enhanced} (ETD), which would be minimized for explicit domain alignment. An alternative solution to the problem is to introduce a domain discriminator that is trained adversarially for implicit domain alignment~\cite{zhang2018collaborative,tzeng2017adversarial,jiang2020implicit,li2019joint}. Domain adaptation has also been introduced in multiple investigations in graph data mining~\cite{wu2020unsupervised,zhang2021adversarial,shen2020network}. For instance, UDA-GCN~\cite{wu2020unsupervised} employs local and global graph neural networks to discover similarity relationships and a domain classifier to align node embeddings on two graphs. AdaGCN~\cite{dai2022graph} leverages both semi-supervised learning and adversarial learning techniques to improve performance. 
Graph Contrastive Learning Network~\cite{wu2022attraction} (GCLN) introduces graph contrastive learning to enhance transfer learning on graphs, whereby similarity relationships are learned from the perspectives of attraction and repulsion forces. In contrast, our \method{} estimates prior distributions and sample subgraphs to surmount potential label shifts, which guide the reliable learning of node representations. 

\subsection{Learning with Label Noise}

Label noise would inevitably be introduced during large-scale data annotation~\cite{zhang2021understanding,tan2021co}. Consequently, learning with label noise has received much attention recently and has been applied to various problems, including image segmentation~\cite{oh2021background} and entity alignment~\cite{pei2020rea}. Potential solutions include modifying loss objectives~\cite{patrini2017making,han2018co,reed2014training,lee2020robust} by incorporating the noise transition matrix or selecting noisy samples with a high degree of confidence. For example, Co-Teaching~\cite{han2018co} introduces two deep neural networks that are independently trained and provide each other with communications about trustworthy examples. 
A small number of studies on learning with label noise on graphs have been conducted recently~\cite{nt2019learning,dai2021nrgnn,qian2023robust,yuan2023learning}. 
NRGNN~\cite{dai2021nrgnn} explores unlabeled nodes alongside labeled nodes to provide more semantic information. PI-GNN~\cite{du2021pi} concentrates on learning pairwise supervised signals based on graph structure, thereby eliminating potential overfitting in graph neural networks. In this paper, we train an extra projector by maximizing mutual information between node representation and labels, then we choose nodes with high inconsistency similarity structures as noisy ones.

\begin{figure*}[ht]
\centering
\includegraphics[width=\textwidth,keepaspectratio=true]{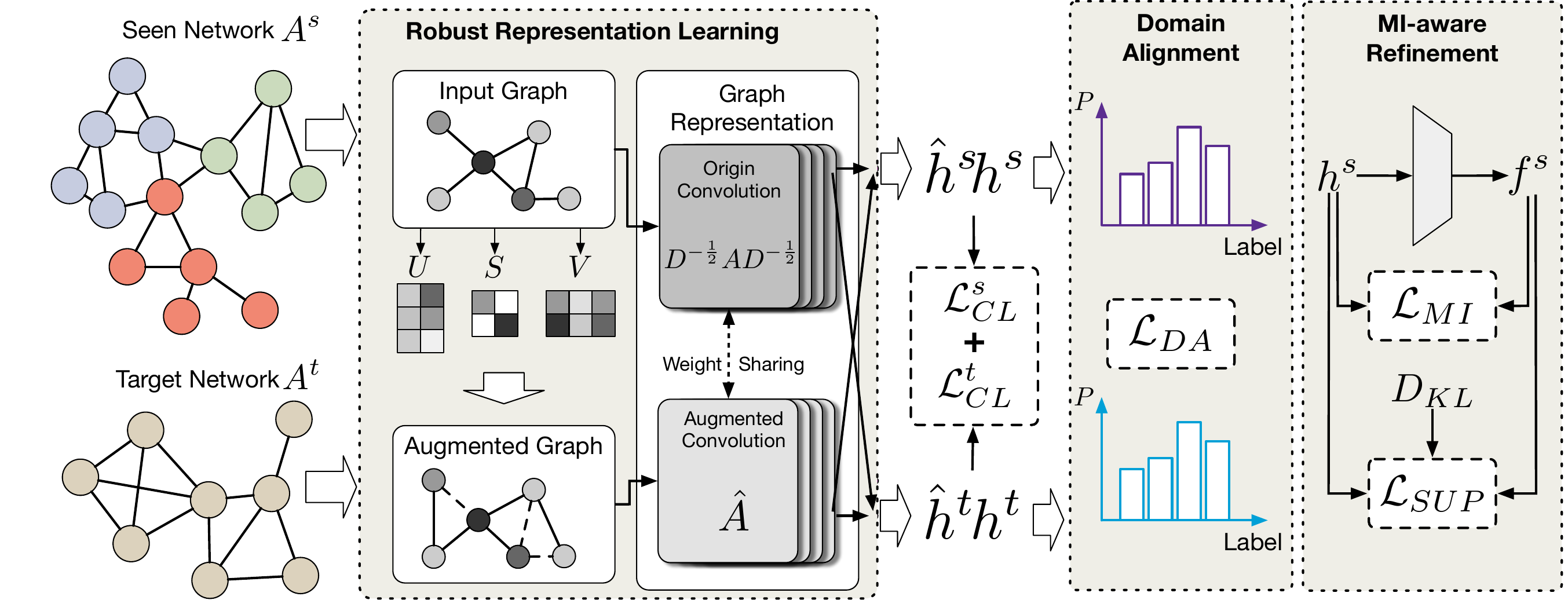}
\caption{Overview of our proposed \method{}. Our \method{} first generates augmented graphs using singular value decomposition and then compares node representations across views with contrastive learning. Moreover, \method{} computes prior label distributions for balanced domain alignment. To clean label noise, \method{} maps node representations into a different space and then measures similarity structures to identify noisy nodes. 
}
\label{fig:framwork}
\end{figure*}

\section{Methodology}

\subsection{Framework Overview}

This paper investigates the problem of graph transfer learning under label noise, which is challenging due to the combined influence of domain shift and label noise. Here, we propose a novel method named \method{}, which is featured three primary components: 
\begin{itemize}[leftmargin=*]
    \item \textbf{Robust Representation Learning.} Considering that label noise and domain shift can introduce potential biases, we employ graph contrastive learning based on singular value decomposition to produce robust representations on two graphs.
    \item \textbf{Balanced Domain Alignment.} To effectively address domain shift alongside prospective label shifts between source and target graphs, we compute a prior distribution to generate subgraphs with balanced label distributions. Subsequently, we apply a domain discriminator to facilitate the alignment of underlying multi-modal distributions in an adversarial manner.
    \item \textbf{Mutual Information-aware Refinement.} We transform node embeddings into an alternative space that maximizes mutual information with label information and identify nodes exhibiting high inconsistency in similarity structures between the two embedding spaces as potentially noisy ones.
\end{itemize}

\subsection{Problem Definition}
Given a labeled source graph consisting of $N^s$ nodes, denoted as $\mathcal{G}^s = (\mathcal{V}^s, \mathcal{E}^s)$, and an unlabeled target graph comprising $N^t$ nodes, denoted as $\mathcal{G}^t = (\mathcal{V}^t, \mathcal{E}^t)$. The attribute matrices for the two graphs are represented by $X^s \in \mathbb{R}^{N^s \times F}$ and $X^t \in \mathbb{R}^{N^t \times F}$, respectively, where $F$ indicates the dimension of node attributes. The adjacency matrices for the two graphs are denoted as $\bm{A}^s \in \{0, 1\}^{N^s \times N^s}$ and $\bm{A}^t \in \{0, 1\}^{N^t \times N^t}$, respectively.
The label for each source node $i^s \in \mathcal{G}^s$ is represented by $y_i^s \in \{1, 2, \cdots, C\}$, where $C$ denotes the number of classes and may contain a certain degree of noise. Our objective is to predict the label $y_i^t$ for each node on the target graph $i^t \in \mathcal{G}^t$ with potential domain shifts.

\subsection{Robust Representation Learning Using Singular Value Decomposition}

To eliminate the confounding effects of domain shift and label noise, we introduce an unsupervised component that generates 
robust node representations without data annotations for every graph. Here, a basic assumption for homophily graphs~\cite{wang2022hagen} is that nodes with similar labels and features tend to be connected, indicating a low rank of similarity matrix based on the ground truth. Therefore, we generate different views using singular value decomposition~\cite{baker2005singular} (SVD) on the adjacency matrix.  
These views are compared with original graphs to formulate contrastive loss, which helps to generate robust representations devoid of data annotation. 

In particular, we first normalize the adjacent matrix and then introduce the SVD algorithm as in \cite{halko2011finding}, which generates an orthonormal matrix with rank $q$. Formally, 
\begin{equation}
{\bm{U}}^s, {\bm{S}}^s, {\bm{V}}^{s\top}=\operatorname{SVD}(\bm{D}^{s^{-\frac{1}{2}}}\bm{A}^s \bm{D}^{s^{-\frac{1}{2}}}, q),
\end{equation}
\begin{equation}\label{eq:aug}
\hat{\bm{A}}^s={\bm{U}}^s {\bm{S}}^s {\bm{V}}^{s,\top},
\end{equation}
where $\bm{D}^s$ is the degree matrix of $\bm{A}^s$ for normalization. ${\bm{U}}^s$, ${\bm{S}}^s$ and ${\bm{V}}^{s,\top}$ are three $p$-rank matrix from the approximate SVD algorithm~\cite{halko2011finding}. 
Then, we leverage graph convolution layers $GC^{l}_\theta$ with parameters $\theta$ to update node representations using our augmented adjacency. In formulation, we have:
\begin{equation}\label{eq:hidden_embed_a}
\begin{aligned} \hat{\bm{H}}^{s,(l)} & =\operatorname{GC}_{\theta}\left(\hat{\bm{A}}^s, \hat{\bm{H}}^{s,(l-1)}\right) \end{aligned}
\end{equation}
where $\hat{\bm{H}}^{s,(l)} \in \mathbb{R}^{N^s \times d}$ is the augmented hidden matrix for source graph at layer $l$, $\bm{W}^{(l)}$ is the weight matrix in graph convolution module and $\sigma(\cdot)$ is an activation function. After stacking $L$ layers, the final node embeddings are written as: 
\begin{equation}
    \hat{\bm{H}}^s  = [\bm{\hat{\bm{h}}}^s_1, \cdots, \bm{\hat{\bm{h}}}^s_N] = GC_\theta(\hat{\bm{A}}^s, \bm{X}^s),
\end{equation}
where $\bm{\hat{\bm{h}}}^s_i$ is the embedding for node $i^s \in \mathcal{G}^s$. Note that $GC_\theta$ can be implemented using vanilla graph convolution~\cite{kipf2017semi} or more flexibly by adopting sophisticated architectures~\cite{wu2019simplifying,wu2020unsupervised}. 

Subsequently, we compare the node embeddings from augmented graphs with those obtained from original graphs, following the principle of graph contrastive learning. Specifically, we produce an original node embedding matrix as:
\begin{equation}\label{eq:hidden_embed}
\begin{aligned} \bm{H}^{s} & =\operatorname{GC}_{\theta}\left(\bm{D}^{s^{-\frac{1}{2}}}\bm{A}^s \bm{D}^{s^{-\frac{1}{2}}} , \bm{X}^{s}\right) \end{aligned}
\end{equation}
where every weight matrix $\bm{W}^{(l)}$ is shared for both node embedding matrices. 
Finally, a graph contrastive learning objective is utilized to generate robust representations as follows:
\begin{equation}
    \mathcal{L}_{CL}^s = - \sum_{i=1}^{N^s} \log \frac{\exp(\bm{h}_i^s \star \hat{\bm{h}}_i^s / \tau ) }{\sum_{j=1}^{N^s} \exp(\bm{h}_i^s \star \hat{\bm{h}}_j^s / \tau ) },
\end{equation}
where $\star$ calculates the cosine similarity of two vectors and $\tau$ is a temperature parameter set to $0.5$ suggested by~\cite{luo2020cimon}. 
Similarly, we generate a contrastive objective for node embeddings in the target graph. Then the node embeddings for original and augmented target graphs are represented as $\bm{h}_i^t$ and $\hat{\bm{h}}_i^t$, which results in $\mathcal{L}_{CL}^t$. The final objective for obtaining robust representations is as follows: 
\begin{equation}
    \mathcal{L}_{CL} = \mathcal{L}_{CL}^s + \mathcal{L}_{CL}^t
\end{equation}

This component does not conduct node representation learning on two single graphs without label information, which can potentially generate robust representations in our scenarios with significant label noise and domain discrepancy issues. 

\subsection{Subgraph Sampling for Balanced Domain Alignment}

Existing methods employ an adversarial domain discriminator~\cite{wu2020unsupervised,li2019joint} to guide domain alignment to address domain discrepancy between source and target graphs. However, potential label shift~\cite{liu2021adversarial} between these two graphs is ignored. In other words, the proportion of each class could vary between the source and target graphs, making it difficult to align underlying multi-modal distributions~\cite{long2018conditional}. Therefore, we propose subgraph sampling, which calculates the prior distribution for each node and then sample subgraphs for balanced label distribution. Then, both hidden features and label distributions are taken into account to identify complex underlying structures for domain alignment. 

In particular, we first introduce an MLP classifier $f_\theta(\cdot):\mathbb{R}^d \rightarrow \mathbb{R}^C $, which maps hidden embedding to label space. The classifier is well-trained by the source graph with the cross-entropy objective:
\begin{equation}\label{eq:sup}
   \mathcal{L}_{SUP} = -\frac{1}{N^s} \sum_{i^s \in \mathcal{G}^s}{{\bm{y}}_{i}^{s}}^T\log\bm{p}^s_{i},
\end{equation}
where $\bm{p}_i^t = f_\theta(\bm{h}_i^t)$ is the predicted distribution and $\bm{y}_{i}^{s}$ is the one-hot label embedding. Then, a pseudo-label can be generated for every target node, i.e., $\hat{y}^t_i = \arg \max_c {\bm{p}_i^t}[c]$. Then, we can estimate the label distribution of the two graphs, i.e., $p(y)$. In formulation, 
\begin{equation}\label{eq:prior}
    p(y)[m] = \frac{\#\{i| y_i^s=m \} + \#\{j| \hat{y}_j^t=m \}} {N^s + N^t},
 \end{equation}
where $p(y)[m]$ denotes the probability of class $m$ in the dataset. Then, we sample nodes from both source and target graphs based on $p(y)$, which generates subgraphs $\tilde{\mathcal{G}}^s$ and $\tilde{\mathcal{G}}^t$ with balanced label distributions. To achieve domain alignment for underlying multi-modal distributions, a domain discriminator conditioned on both hidden features and labels is introduced, which guides feature learning in an adversarial manner. In formulation, we utilize an MLP to formalize $D_\phi$ and the objective is stated as:
 \begin{equation}
    \mathcal{L}_{DA}(\theta, \phi) = - \sum_{i\in \hat{\mathcal{G}}^s} \log D_\phi([\bm{h}_i^s, \bm{y}_i^s])  - \sum_{j \in \hat{\mathcal{G}}^t }\log (1-D_\phi([\bm{h}_j^t, \bm{p}_j^t])),
\end{equation}
where hidden information and label information is concatenated as the input of $D_\phi$. On the one side, the objective is minimized with regard to $\phi$ to distinguish two domains utilizing all available resources. On the other side, the objective is maximized with respect to $\theta$ to confound the domain discriminator, thereby decreasing the domain disparity of multi-modal distributions in the hidden space. By employing subgraph sampling, the influence of label shifts is mitigated, reducing bias when aligning multimodal distributions.

\subsection{Mutual information-aware Refinement}

Nevertheless, label noise could still be detrimental to the optimization procedure. One potential solution is to identify noisy samples~\cite{han2018co,lee2020robust}. Existing methods concentrate on independent and identically distributed (i.i.d) data~\cite{han2018co,reed2014training,lee2020robust}, which makes their applications to graph data challenging. To this end, we introduce a mutual information-based refinement module that projects node representations to another embedding space by maximizing mutual information with labels. Since graph structures are not involved, noisy nodes can be significantly influenced. Therefore, nodes with high inconsistency of similarity structures between two spaces are subsequently considered to be noisy. 

In particular, we first introduce an MLP $g_\phi(\cdot)$, which maps node embeddings into another space, i.e., $\bm{f}^s_i = g_\phi(\bm{h}_i^s)$. Here, the mutual information between $\bm{f}^s_i$ and $\bm{y}_i^s$ is maximized. Afterwards, we follow MINE~\cite{belghazi2018mutual} to calculate the lower bound of $I(\bm{f}^s_i, \bm{y}_i^s)$ by introducing an MI estimator $T$. Then, the objective can be formulated as: 
\begin{equation}
   \mathcal{L}_{MI} = \frac{1}{N^s}\sum_{i=1}^{N^s} \left[T(\bm{f}^s_i, \bm{y}_i^s)\right]-\log(\frac{1}{N^{s^2}}\sum_{i=1}^{N^s}\sum_{j=1}^{N^s}\left[e^{T(\bm{f}^s_i, \bm{y}_j^s)}\right]).
\end{equation}

\begin{algorithm}[t]
\caption{Training Algorithm of \method{}}
\label{alg1}
\begin{algorithmic}[1]
\REQUIRE Source graph $\mathcal{G}^s$; Target graph $\mathcal{G}^t$; Percentile $\alpha$ for label refinement; SVD order $q$;\\
\ENSURE Parameters $\theta$ and $\phi$;
\STATE Warm up the network by minimize $\mathcal{L}_{SUP} +  \mathcal{L}_{CL}$;
\STATE Generate the augmented adjacent matrix using Eqn. \ref{eq:aug};
\FOR{$c=1,2,\cdots,C$}
\STATE Obtain the clean node set using Eqn. \ref{eq:clean};
\STATE Calculate $p(y)$ using Eqn. \ref{eq:prior};
\STATE Generate subgraph $\mathcal{G}^s$ and $\mathcal{G}^t$ based on $p(y)$;
\REPEAT
\STATE Generate hidden embeddings for both original and augmented graphs using Eqns. \ref{eq:hidden_embed_a} and \ref{eq:hidden_embed}, respectively. 
\STATE Update parameters of $G(\cdot)$ and $D(\cdot)$ using Eq. \ref{eq:final_loss};
\UNTIL Convergence
\ENDFOR
\end{algorithmic}
\end{algorithm}

Since the absolute positions of all these deep features would change after the projection, we compare the similarity structures rather than positions to identify noisy nodes. 
Here, we define similarity structures by introducing a range of anchor nodes, i.e., $\mathbb{S} = \{i_1^s, \cdots, i_K^s\}\in \mathcal{G}^s$ and the similarity distribution between each node and anchor nodes in the hidden space can be formulated as: 
\begin{equation}
\bm{w}_{i}^{s}[k]=\frac{\exp \left( \bm{h}^s_i \star \bm{h}^s_{i_k} / \tau\right)}{\sum_{k'=1}^{K} \exp \left( \bm{h}^s_i \star \bm{h}^s_{i_{k'}} / \tau\right)}.
\end{equation}
Similarly, the similarity distribution within the newly established embedding space can be described as follows:
\begin{equation}
\bm{r}_{i}^{s}[k]=\frac{\exp \left( \bm{f}^s_i \star \bm{f}^s_{i_k} / \tau\right)}{\sum_{k'=1}^{K} \exp \left( \bm{f}^s_i \star \bm{f}^s_{i_{k'}} / \tau\right)},
\end{equation}
Subsequently, we measure the KL divergence between the two distributions obtained above as inconsistency scores:
\begin{equation}
    d_i = D_{\mathrm{KL}}(\bm{w}_{i}^{s} \| \bm{r}_{i}^{s})
\end{equation}
Given a threshold $\mu$, we identify samples with large inconsistency scores and systematically eliminate them in supervised training. The clean node set is written as follows:
\begin{equation}\label{eq:clean}
C = \{i^s \in \mathcal{G}^s | d_i < \mu(\alpha)\}.  
\end{equation}
where $\mu(\alpha)$ is determined by the $\alpha$ percentile threshold of divergence in the whole source graph. 
In the end, we revise the supervised learning loss in Eqn. \ref{eq:sup} as follows:
\begin{equation}\label{eq:sup_update}
   \mathcal{L}_{SUP} = -\frac{1}{|C|} \sum_{i^s\in C }{{\bm{y}}_{i}^{s}}^T\log\bm{p}^s_{i}.
\end{equation}

\subsection{Framework Summarization}
In a nutshell, the whole loss objective for our proposed \method{} can be succinctly expressed as follows:
\begin{equation}\label{eq:final_loss}
    \min_{\theta} \max_{\phi} \mathcal{L} = \mathcal{L}_{SUP} +  \mathcal{L}_{CL} - \mathcal{L}_{DA} +  \mathcal{L}_{MI}.
\end{equation}
where $\phi$ is the parameters of $D(\cdot)$ and $\theta$ is for graph neural network. Here, we fix $\phi$ to minimize $\mathcal{L}_{DA}$ and then minimize $\mathcal{L}$ with respect to $\theta$ with fixed $\phi$, which follows the paradigm of adversarial learning. We first warm up our \method{} using contrastive learning loss and supervised loss. The algorithm is summarized in Algorithm \ref{alg1}.

\section{Experiment}
\label{sec::experiment}

In this section, we showcase the experimental findings to establish the potency of \method{}. To start, we will provide an overview of the datasets, baselines, and configurations, followed by a presentation of the results and a comprehensive analysis. We aim to answer the following five research questions in detail. 

\noindent\textbf{RQ1: }How does the proposed \method{} perform on real-world datasets compared with current state-of-art methods?

\noindent\textbf{RQ2: }Is our \method{} robust against different levels of label noise?

\noindent\textbf{RQ3: }How do the hyper-parameters in \method{} affect the cross-domain classification performance ?

\noindent\textbf{RQ4: }What is the role of the main components in our proposed \method{} and how do they impact the overall performance? 

\noindent\textbf{RQ5: }Is there any additional analysis that can confirm
the superiority of \method{} compared to baseline models?

\begin{table}[!t]
	\centering
	\tabcolsep=4pt
	\caption{Dataset Statistics. A, C, and D represent ACM, Citation, and DBLP, respectively. The number in the front of dataset name indicates the specific group. Cross domain tasks are assigned within the same group.}
	\begin{tabular}{c  c  c  c  c  c} 
	
	\toprule

	  Dataset  & $\#$ Label & $\#$ Feature & $\#$ Node & $\#$ Edge & Origin Data  \\
	\midrule
	1A &  6  &  7537  &  5578  & 7341 & ACMv9 \\
        1D &  6  &  7537  &  11135  & 7410 & DBLPv8 \\
	\midrule
        2A &  5  &  6775  &  9369  & 15602 & ACMv9 \\
        2C &  5  &  6775  &  8935  &  15113& Citationv1 \\
        2D &  5  &  6775  &  5484  & 8130  & DBLPv7 \\


	\bottomrule
	\end{tabular}
	\label{tab:dataset}
\end{table}

{
\begin{table*}[ht]
    \centering
    \caption{Summary of accuracy on eight cross-domain classification tasks with Pair Noise. The best performance is highlighted in boldface. Our proposed method \method{} outperforms all the baseline methods in most cases.}\label{tab:main_results-pair}
    \resizebox{0.9\textwidth}{!}{ %
    \begin{tabular}{c|cc c cc c c cc}
    \toprule\midrule

    Dataset & MLP & DeepWalk & GraphSAGE & LINE & GCN & DGRL & AdaGCN & UDAGCN & Ours \\
    \midrule

1A→D& 38.12$\pm$1.9& 28.93$\pm$3.2& 35.60$\pm$9.6& 35.20$\pm$1.1& 56.88$\pm$3.2& 36.87$\pm$0.9& 57.94$\pm$5.8& 67.54$\pm$11.2& \textbf{79.31$\pm$6.4}\\
1D→A& 48.71$\pm$2.9& 37.44$\pm$1.5& 36.52$\pm$2.2& 35.09$\pm$2.6& 65.60$\pm$1.1& 44.52$\pm$4.8& 62.59$\pm$2.8& 65.81$\pm$2.7& \textbf{71.22$\pm$1.4}\\

\midrule

2A→D& 43.11$\pm$1.4& 15.98$\pm$3.9& 52.96$\pm$3.4& 18.57$\pm$2.1& 52.10$\pm$2.4& 40.32$\pm$1.6& 51.23$\pm$6.1& 52.96$\pm$7.7& \textbf{54.54$\pm$4.4}\\
2D→A& 34.90$\pm$2.0& 30.28$\pm$2.4& 36.42$\pm$4.3& 31.67$\pm$1.8& 34.08$\pm$2.3& 32.86$\pm$3.8& 36.93$\pm$5.4& 37.40$\pm$7.2& \textbf{56.39$\pm$4.2}\\
2A→C& 39.58$\pm$1.7& 23.19$\pm$5.8& 48.89$\pm$5.7& 24.64$\pm$5.7& 46.30$\pm$5.6& 36.26$\pm$2.4& 47.01$\pm$7.5& 47.11$\pm$9.1& \textbf{64.88$\pm$6.0}\\
2C→A& 38.69$\pm$2.0& 19.43$\pm$3.9& 45.77$\pm$5.0& 22.54$\pm$5.3& 47.10$\pm$4.4& 35.34$\pm$2.3& 43.65$\pm$4.7& 43.36$\pm$6.8& \textbf{60.79$\pm$4.7}\\
2C→D& 44.74$\pm$1.7& 13.29$\pm$1.7& 57.26$\pm$2.0& 15.18$\pm$1.6& 57.69$\pm$2.7& 41.95$\pm$1.0& 55.63$\pm$4.4& \textbf{57.20$\pm$6.1}& 56.96$\pm$3.7\\
2D→C& 35.32$\pm$2.3& 24.36$\pm$5.1& 38.35$\pm$6.2& 24.76$\pm$1.1& 35.67$\pm$4.5& 37.46$\pm$2.5& 41.36$\pm$9.1& 41.58$\pm$11.3& \textbf{59.93$\pm$4.1}\\

    \midrule\bottomrule
    \end{tabular}
    }
\end{table*}
}

\begin{figure*}[ht]
\centering
\includegraphics[width=\textwidth,keepaspectratio=true]{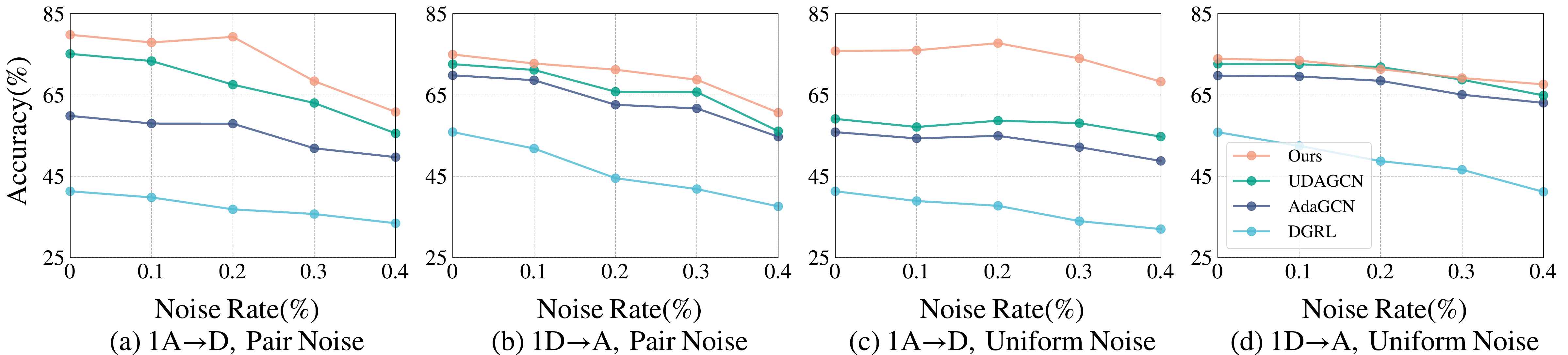}
\caption{
Accuracy on 1A→D and 1D→A with Pair Noise and Uniform Noise on various levels of label noise setting.
}
\label{fig:main_results}
\end{figure*}
\subsection{Experimental Setup}
\noindent\textbf{Benchmark Datasets.}
In this experiment, we diligently assess the performance of our proposed model by examining it on three real-world networks sourced from ArnetMiner~\cite{tang2008arnetminer}. The networks are constructed based on paper citation information obtained from various databases, namely DBLP, ACM, and Microsoft Academic Graph. 
Detailed dataset information is presented in Table~\ref{tab:dataset}.

Since all labels in the datasets are clean, following~\cite{patrini2017making,yu2019does,dai2021nrgnn}, we employ two different disturbance methods to add noise to the training labels of the original datasets. (i) Uniform Noise: The label of each node is randomly switched to another class with a certain probability. (ii) Pair Noise: The label of each node is changed exclusively to its most similar class with a certain probability. In both two methods, the probability is a parameter set to 20\% by default. To appropriately choose the best model and assess its performance, we split the dataset into training, testing, and validation sets, maintaining a 7:2:1 proportion for each respective set.

\noindent\textbf{Evaluation Protocols.}
To evaluate our algorithm for cross-domain node classification, we utilized two distinct sets of data partitioning. The first set comprises DBLPv7, ACMv9, and Citationv1, which have been commonly used in previous domain adaptation studies~\cite{dai2022graph}. The second set consists of more recent data, DBLPv8, and ACMv9, with slightly different feature preprocessing~\cite{wu2020unsupervised}. 

To simplify the notation, we referred to the dataset groups using their corresponding numbers and initials. Specifically, we designated the group with DBLPv8 and ACMv9 as Group 1, and the older group with DBLPv7, ACMv9, and Citationv1 as Group 2. Therefore, we used the notation 2C to refer to Citationv1 from Group 2. We conducted multi-label classification on these three network domains, performing eight transfer learning tasks: 1A→D, 1D→A, 2A→D, 2D→A, 2A→C, 2C→A, 2C→D, and 2D→C.

We assess all models by conducting a grid search within the hyperparameter space and subsequently present the optimal results for each respective approach. For each task, we perform ten trials with all models by varying the random seed and report the accuracy in percentage with standard deviation for comparison.

\noindent\textbf{Baseline Methods.}
We compare \method{} with state-of-the-art models. Note that for fairly evaluating our model and highlight its performance, our approach is tested against both advanced single-domain node classification models and cross-domain models. 
The single-domain models consist of methods that ignore graph information (MLP), models that use neighborhood information and graph structure (DeepWalk~\cite{perozzi2014deepwalk}, LINE~\cite{tang2015line}), and graph neural network models (GraphSAGE~\cite{hamilton2017inductive}, GCN~\cite{kipf2017semi}). The cross-domain models include DGRL~\cite{ganin2016domain}, AdaGCN~\cite{dai2022graph}, and UDAGCN~\cite{wu2020unsupervised}, which employs gradient reverse layer (GRL) for domain classification. 
DGRL utilizes an MLP as a feature generator while AdaGCN and UDAGCN adopt GCN architecture to focus on graph representation learning.

{
\begin{table*}[ht]
    \centering
    \caption{Summary of accuracy on eight cross-domain classification tasks with Uniform Noise. The best performance is highlighted in boldface. Our proposed method \method{} outperforms all the baseline methods in most cases.}\label{tab:main_results-uniform}
    \resizebox{0.9\textwidth}{!}{ %
    \begin{tabular}{c|cc c cc c c cc}
    \toprule\midrule

    Dataset & MLP & DeepWalk & GraphSAGE & LINE & GCN & DGRL & AdaGCN & UDAGCN & Ours \\
    \midrule

1A→D& 37.77$\pm$1.7& 27.19$\pm$3.8& 30.22$\pm$4.1& 32.83$\pm$1.7& 49.32$\pm$3.8& 37.73$\pm$1.4& 54.95$\pm$7.2& 58.68$\pm$9.9& \textbf{77.75$\pm$5.3}\\
1D→A& 50.56$\pm$3.0& 38.70$\pm$1.2& 43.92$\pm$4.8& 36.05$\pm$2.0& 68.29$\pm$1.3& 48.72$\pm$2.8& 68.48$\pm$1.7& \textbf{71.88$\pm$1.9}& 71.33$\pm$2.4\\

\midrule

2A→D& 45.25$\pm$1.7& 13.68$\pm$4.9& 55.18$\pm$4.8& 14.82$\pm$2.0& 41.88$\pm$6.1& 41.37$\pm$0.7& 48.14$\pm$7.1& 48.05$\pm$9.1& \textbf{63.12$\pm$4.4}\\
2D→A& 35.31$\pm$2.0& 31.93$\pm$2.4& 36.76$\pm$3.5& 32.44$\pm$1.9& 31.58$\pm$2.7& 34.12$\pm$2.8& 35.26$\pm$7.5& 33.34$\pm$8.6& \textbf{62.63$\pm$2.5}\\
2A→C& 40.54$\pm$1.7& 26.96$\pm$5.9& 50.93$\pm$5.7& 28.02$\pm$5.8& 33.62$\pm$5.9& 40.89$\pm$2.3& 39.77$\pm$7.1& 42.03$\pm$9.8& \textbf{68.63$\pm$4.6}\\

2C→A& 39.98$\pm$3.3& 17.86$\pm$5.1& 49.59$\pm$7.3& 24.08$\pm$2.4& 38.73$\pm$8.2& 39.98$\pm$1.6& 43.84$\pm$4.4& 42.10$\pm$6.7& \textbf{65.81$\pm$3.3}\\
2C→D& 47.10$\pm$4.7& 11.62$\pm$1.3& 62.14$\pm$4.6& 13.49$\pm$1.3& 53.98$\pm$2.6& 43.71$\pm$1.5& 56.74$\pm$5.5& 55.45$\pm$9.9& \textbf{66.73$\pm$4.1}\\
2D→C& 36.08$\pm$2.5& 24.64$\pm$6.1& 38.98$\pm$3.6& 26.47$\pm$0.9& 29.51$\pm$4.4& 36.08$\pm$1.9& 39.58$\pm$4.4& 35.46$\pm$10.6& \textbf{68.18$\pm$4.3}\\


    \midrule\bottomrule
    \end{tabular}
    }
\end{table*}
}

\begin{figure*}[ht]
\centering
\includegraphics[width=\textwidth,keepaspectratio=true]{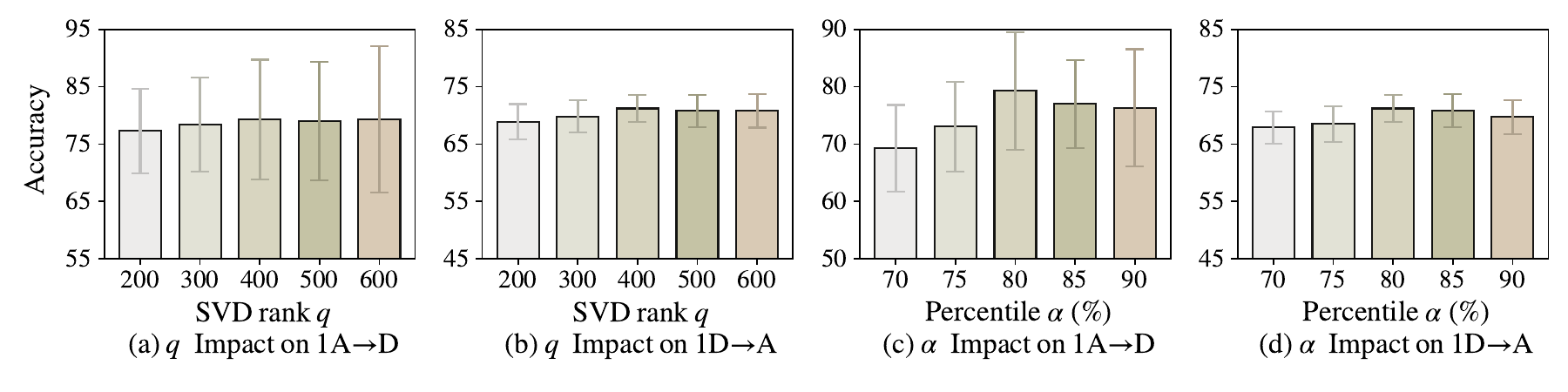}
\caption{Parameter Sensitivity Analysis on 1A→D and 1D→A tasks with Pair Noise. Bar charts and error bars depict accuracy and corresponding 80\% confidence intervals. (a)(b) display performance on different SVD rank $q$ and (c)(d) shows performance varying percentile $\alpha$ of KL divergence threshold.
}
\vspace{-0.2cm}
\label{fig:para_sensi}
\end{figure*}

\subsection{Performance Comparison (RQ1)}

Table~\ref{tab:main_results-pair} and Table~\ref{tab:main_results-uniform} illustrate the accuracy in percentage with standard deviations of \method{} and baseline models across all tasks, considering the Pair and Uniform Noise with default noise rate of 20\%. The results reveal the following insights:

\begin{itemize}[leftmargin=*]
\item Models employing traditional encoders, such as MLP and DGRL, show suboptimal performance due to their inability to capture the complex relationships within  graph structure. This underscores the necessity of leveraging the inductive bias of graph structures.

\item Graph-based approaches, such as DeepWalk, LINE, GraphSAGE, and GCN, demonstrate enhanced performance in comparison to traditional methods, highlighting the benefits of leveraging both local graph structures and node features for superior representation. Furthermore, GraphSAGE and GCN stand out due to their graph convolution capabilities, which allow models to preserve structural features on multiple scales.

\item Multi-task learning models, such as DGRL and AdaGCN, outperform single-task models, indicating the benefits of jointly learning from multiple tasks. Nevertheless, these models still underperform compared to our proposed \method{}, as they are notably impacted by noisy labels. This underscores the challenge of multitask node classification in the presence of label noise.

\item Our proposed \method{} consistently outperforms all baseline models across the majority of tasks, showcasing its exceptional ability to capture the complex relationships within multi-domain graph-structured data and reduce the impact of noise. Specifically, \method{} exceeds the best-performing baseline by 28.60\% and 25.87\% on 2D→C and 2D→A, respectively. Additionally, \method{} displays a relatively smaller standard deviation compared to other graph-based multi-task models, signifying a more robust and stable performance across diverse training scenarios.
\end{itemize}

\subsection{Effects of Different Noisy Rates (RQ2)}

To demonstrate the effectiveness of the \method{} model in handling different levels of label noise, we evaluate its performance with noise rates ranging from $\{0, 10, 20, 30, 40\}\%$. We also compare our method with competitive baselines.
We evaluate our method and baselines on the tasks of 1A→D and 1D→A with both Pair and Uniform Noise. The mean performance across ten trials is demonstrated in Figure~\ref{fig:main_results}. Our observations from the figure are as follows:

\begin{itemize}[leftmargin=*]
\item Even in the presence of low or higher label noise, the proposed \method{}  surpasses the other models.
This is attributed to the robust representation learning of Singular Value Decomposition based contrastive learning, which enhances GNN message-passing process. Also, balanced domain alignment contributes to performance by reducing classification bias with or without noise.
\item With the increase in label noise levels, a significant decline in the performance of all baselines is observed. Although the performance of \method{} is also affected, it still maintains a lead over the baselines, which exhibits greater resilience in the face of label noise, indicating the efficacy of the proposed method in managing noisy labels by mutual information-aware refinement.
\end{itemize}

\subsection{Parameter Sensitivity Analysis (RQ3)}

In this section, we study the impact of two hyper-parameters in our \method{}: the SVD rank $q$ in the SVD-based graph contrastive learning module and the percentile threshold $\alpha$ in the mutual information-aware refinement procedure, which measures whether inconsistency scores of each node are significant enough to be retained in supervised training. We evaluate the model's performance with different values of $q$ and $\alpha$ and report the results on a chosen dataset.

\noindent\textbf{Impact of SVD rank $q$}: We conducted experiments by varying the SVD rank $q$ in the SVD-based graph contrastive learning module from 200 to 600 and evaluated the performance on transfer learning tasks 1A→D and 1D→A. The results are presented in Figure~\ref{fig:para_sensi}. We observe that the performance is relatively stable across different values of $q$, which indicates that our \method{} model is robust to changes in the SVD rank $q$. However, the variations in performance may increase with a larger value of $q$. Additionally, it can be noticed that the best performance was achieved when setting $q$ to 400. So we kept $q=400$ by default in other experiments.

\noindent\textbf{Impact of $\alpha$}: We conducted experiments to investigate the impact of the threshold in the mutual information-aware refinement module. We varied the percentile of threshold, $\alpha$, over $\{ 70, 75, 80, 85, 90 \} \%$ and plotted the results in Figure~\ref{fig:para_sensi}. It can be observed that the classification performance improves gradually when $\alpha$ increases from 70\% to 80\%. This improvement is mainly due to the fact that lower threshold values set a more strict criterion for inclusion, leading to the removal of more training samples. However, we also found that excessively large threshold values could have a negative impact on performance, as they tend to increase the number of noisy samples that interfere with the optimization procedure. 
So it seems that the best $\alpha$ value should be set between 75\% and 85\%. Our experiment suggests that this range provides optimal results in different tasks.

\subsection{Ablation Study (RQ4)}

To understand the effectiveness of each component in our proposed \method{}, we conduct an ablation study by comparing different variants of \method{}. We investigate the impact of the SVD-based graph contrastive learning, balanced domain alignment module, and mutual information-aware refinement procedure for denoising. The results of the ablation study are presented in Table~\ref{tab:ablation}, and the following \method{} variants are designed for comparison:

\begin{itemize}[leftmargin=*]
\item \method{} w/o G: A variant of \method{} with the SVD-based graph contrastive learning component removed.
\item \method{} w/o B: A variant of \method{} with the subgraph sampling for the balanced domain alignment module removed.
\item \method{} w/o M: A variant of \method{} with the mutual information-aware refinement procedure removed.
\end{itemize}

{
\begin{table}[t]
    \centering
    \caption{Comparisons between our \method{} and its variants in different settings
    with Pair Noise. }\label{tab:ablation}
    \resizebox{0.48\textwidth}{!}{ %
    \begin{tabular}{c|ccc c}
    \toprule\midrule
    
    Dataset & \method{} w/o G & \method{} w/o B &	\method{} w/o M &	\method{} \\
    \midrule

2A→D& 49.44$\pm$9.5& 50.57$\pm$5.1& 53.14$\pm$3.7& \textbf{54.54$\pm$4.4}\\
2D→A& 40.83$\pm$8.3& 38.91$\pm$4.6& 54.07$\pm$3.9& \textbf{56.39$\pm$4.2}\\
2A→C& 47.46$\pm$9.3& 47.25$\pm$6.3& 56.73$\pm$5.0& \textbf{64.88$\pm$6.0}\\
2C→A& 49.00$\pm$6.1& 45.29$\pm$5.0& 49.72$\pm$3.2& \textbf{60.79$\pm$4.7}\\
2C→D& 55.21$\pm$8.9& 56.16$\pm$3.7& \textbf{57.36$\pm$3.9}& 56.96$\pm$3.7\\
2D→C& 47.80$\pm$10.8& 44.00$\pm$3.8& 57.92$\pm$4.2& \textbf{59.93$\pm$4.1}\\

    \midrule\bottomrule
    
    \end{tabular}
    }
\end{table}
}

\noindent\textbf{Impact of SVD-based Graph Contrastive Learning}: Comparing \method{} with \method{} w/o G helps us understand the effectiveness of the SVD-based graph contrastive learning component. The results show that \method{} outperforms \method{} w/o G, which indicates that SVD-based graph contrastive learning benefits the model by extracting robust node representation bypassing the label noise.

\noindent\textbf{Impact of Balanced Domain Alignment}: To verify the effectiveness of the balanced domain alignment module, we compare \method{} and \method{} w/o B. From the results, we can observe that \method{} performs significantly better than \method{} w/o B, indicating that the subgraph sampling with balanced domain alignment effectively addresses the problem of domain shift along with label shifts.

\noindent\textbf{Impact of Mutual Information-aware Refinement}: We compare \method{} with \method{} w/o M to investigate the effectiveness of the mutual information-aware refinement procedure, which identifies noisy nodes with high inconsistency.  The results show that \method{} achieves better performance than \method{} w/o M in most datasets, demonstrating the importance of the mutual information-aware refinement procedure in improving the model's resilience to noise.

\subsection{Visualization (RQ5)}
\begin{figure}[t]
\centering
\includegraphics[width=0.48\textwidth,keepaspectratio=true]{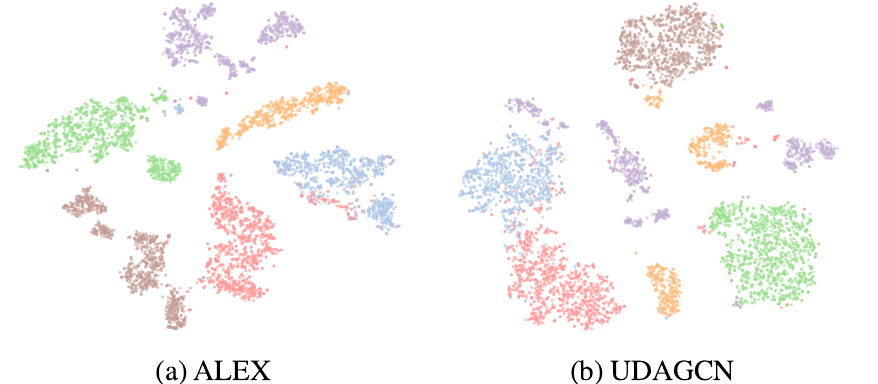}
\caption{Visualization of vector embedding with Dimensionality Reduction by t-SNE. Each point represents a node and the colors stand for different labels.
}
\label{fig:tsne}
\end{figure}

Furthermore, we investigate the representational ability by visualizing the learned embedding, projected with T-distributed Stochastic Neighbor Embedding ($t$-SNE), in a 2-D space. Fig~\ref{fig:tsne} compares the visualization results between our \method{} and best baseline UDAGCN for the 1A→D task with Pair Noise and default label noise rate. We can observe that Embeddings generated by our \method{} exhibit more clear and distinct boundary lines between different classes compared to the best baseline.
This demonstrates that our \method{} can produce more meaningful graph embedding than other approaches, which explains our better classification performance.

\section{Conclusion}
\label{sec::conclusion}

This paper studies the problem of graph transfer learning under label noise, which aims to transfer knowledge from a noisy source graph to an unlabeled target graph. We propose a novel method named \method{} to solve the problem, which first generates robust node representations using graph contrastive learning. Then, we calculate the prior distribution for subgraph sampling for balanced domain alignment. Finally, we leverage the mutual information maximum to identify noisy samples which overfit noisy labels. Extensive experiments on various datasets substantiate the superiority of the proposed \method{}. 
In future work, we would extend our \method{} model to address more realistic challenges, including label imbalance, and zero-shot learning on graphs. 

\section*{Acknowledgements}
This paper is partially supported by the National Natural Science Foundation of China with Grant No. 62276002 as well as the China Postdoctoral Science Foundation with Grant No. 2023M730057.

\bibliographystyle{ACM-Reference-Format}
\bibliography{7_rec}
\balance



\end{document}